# Evaluating large language models in medical applications: a survey

**Authors:**

Xiaolan Chen[1], Jiayang Xiang[4], Shanfu Lu[5], Yexin Liu[6], Mingguang He[1,2,3], Danli Shi[1,2*]

**Affiliations:**

[1] School of Optometry, The Hong Kong Polytechnic University, Kowloon, Hong Kong.

[2] Research Centre for SHARP Vision (RCSV), The Hong Kong Polytechnic University, Kowloon, Hong Kong.

[3] Centre for Eye and Vision Research (CEVR), 17W Hong Kong Science Park, Hong Kong.

[4] Department of Ophthalmology, Renji Hospital, School of Medicine, Shanghai Jiao Tong University, Shanghai 200127, China.

[5] Perception Vision Medical Technologies Co. Ltd., Guangzhou 510530, China.

[6] AI Thrust, The Hong Kong University of Science and Technology, Guangzhou, China.

**Correspondence:**

***Dr Danli Shi,** MD, PhD.

School of Optometry, The Hong Kong Polytechnic University, Kowloon, Hong Kong SAR, China.

Email: danli.shi@polyu.edu.hk.

**ABSTRACT**

Large language models (LLMs) have emerged as powerful tools with transformative potential across numerous domains, including healthcare and medicine. In the medical domain, LLMs hold promise for tasks ranging from clinical decision support to patient education. However, evaluating the performance of LLMs in medical contexts presents unique challenges due to the complex and critical nature of medical information. This paper provides a comprehensive overview of the landscape of medical LLM evaluation, synthesizing insights from existing studies and highlighting evaluation data sources, task scenarios, and evaluation methods. Additionally, it identifies key challenges and opportunities in medical LLM evaluation, emphasizing the need for continued research and innovation to ensure the responsible integration of LLMs into clinical practice.

## INTRODUCTION

In recent years, the emergence of large language models (LLMs) has catalyzed transformative advancements across diverse domains, spanning from natural language understanding to content generation. These models, with expanding capabilities, are increasingly finding integration into a wide spectrum of applications.[1-3] Notably, researchers have begun exploring their potential in the medical field,[4,5] from aiding clinical decision-making to enhancing patient education and engagement.[6]

However, the inherent limitations of LLMs, particularly in tasks like interpreting medical images or grasping clinical context, necessitate a shift towards multi-modal LLMs. These models, integrating text with images and clinical data, enable a more comprehensive analysis and interpretation, thereby improving various aspects of healthcare delivery, including diagnosis,[7,8] treatment planning,[9] report generation,[10] and patient education in healthcare settings.[11]

This trend highlights the importance of evaluating and benchmarking large language models tailored specifically for medical applications.[12] Such evaluations ensure that these models meet the rigorous demands of the healthcare domain in terms of accuracy, reliability, and interpretability.[13] By establishing standardized evaluation criteria and benchmarks,[14] researchers and developers can objectively assess the performance of medical LLMs, driving advancements in healthcare AI and ultimately enhancing patient care and clinical outcomes.[13]

This paper aims to provide a comprehensive overview of the landscape of medical LLM evaluation (**Figure 1**), synthesizing insights from existing studies and addressing key challenges and opportunities. Specifically, we first provide a holistic perspective on their capabilities and challenges through exploration of data sources and task scenarios, paving the way for targeted improvements and innovations. Second, we synthesize diverse evaluation methodologies employed in medical LLM evaluation, ranging from objective accuracy metrics to more nuanced human-centric evaluations. By categorizing and analyzing these methodologies, this article offers insights into the strengths and limitations of different evaluation approaches, guiding future research directions. Third, by identifying key challenges and opportunities in medical LLM evaluation, this paper underscores the need for continued research and innovation in the field. By providing a coherent understanding of the latest techniques in medical LLM evaluation, we hope to offer conceptual advancements to facilitate the responsible integration of LLMs into medical practice.

## DATA SOURCE

Before applying LLMs in medical applications, a thorough evaluation of the LLMs is crucial. The complexity and diversity of medical data pose a significant challenge when constructing appropriate test sets. Currently, such datasets can be broadly divided into two main categories: existing medical resources and manually curated question sets.

### Existing Medical Resources

Medical examinations, designed to assess the competency of healthcare professionals, offer readily available benchmark datasets. These exams, crafted through generations of expertise and accompanied by standardized answers, provide a substantial volume of validated material for evaluating LLMs. Diverse medical exams from different countries have been used to assess LLMs' general medical knowledge capabilities, including the United States Medical Licensing Examination (USMLE),[15,16] the Medical College Admission Test (MCAT),[17] the National Medical Licensing Examination (NMLE) in China,[18] the National Pharmacist Licensing Examination (NPLE) in China, National Nurse Licensing Examination in China,[19] Chinese Master's Degree Entrance Examination[20] and more. For more focused assessment in specific sub-specialties in medicine, exams such as the Ophthalmic Knowledge Assessment Program (OKAP) examination,[21] the Basic Science and Clinical Science Self-Assessment Program,[22] the oral and written board examinations for the American Board of Neurological Surgery (ABNS),[23] Otolaryngology-Head and Neck Surgery Certification Examinations,[24] the Royal College of General Practitioners Applied Knowledge Test,[25] and others are utilized. These datasets provide a wealth of data aligned with specialty knowledge and practices to the assessment of the depth and breadth of knowledge possessed by LLMs.

In addition to examinations, medical literature serves as an important knowledge repository, including peer-reviewed journal articles and conference papers.[26,27] These databases offer cutting-edge medical insights and research findings, contributing to assessing whether the LLM is adept at rapidly updating medical knowledge.

**Manually Curated Questions**

While the range of exam questions and academic materials is extensive, their ability to reflect the dynamic capabilities required for real-world interactions remains limited. Therefore, some studies have turned to the use of real-world data manually created or collected by healthcare professionals for evaluation. For instance, Singhal et al.[4] proposed MultiMedQA as the evaluation datasets for evaluating their developed models. Other studies leveraged real-world interactions and discussions from medical forums and social media to evaluate the conversational and consultative skills of LLM.[28-30] Medical images, such as X-rays, MRI, CT in radiology,[31] fundus photographs, FFA and OCT images in ophthalmology,[32] are essential for tasks like disease diagnosis and image analysis. These clinically derived image data often come with expert medical reports and are crucial resources for constructing multimodal datasets, thus facilitating the testing of LLMs' ability to handle complex visual and textual information.

However, collecting appropriate data can be challenging due to scarce resources in certain disciplines, as well as ethical considerations and data privacy issues. In response, some studies have turned to expert-crafted questions carefully formulated based on clinical expertise.[33-35] For example, Marshall et al.[36] have constructed datasets centered around symptoms, examinations, and treatments of uveitis to evaluate ChatGPT's proficiency in handling specialized content. Another study involved collaboration with a panel of eight board-certified clinicians and two healthcare practitioners, who generated a dataset of 314 clinical questions spanning nine medical specialties, to assess the performance of LLMs.[37] Although these approaches may result in datasets with a limited number of questions, they provide highly specialized and practical insights into LLMs, reflecting the nuanced understanding required in real-world clinical practice. Additionally, the manual creation of test questions guarantees their exclusion from the training data, ensuring its integrity and preventing contamination.

## TASK SCENARIOS

With the rapid development of LLMs, they have demonstrated extensive application prospects in the medical field. The evaluation protocol of the models varies depending on the specific nature of each task. This section aims to review the recent advancements of evaluating LLMs in different medical scenarios, particularly focusing on tasks such as medical closed-ended tasks, open-ended medical question answering (QA), image processing and real-world medical scenarios. **Figure 2** summarizes the data source and evaluation dimension related to different task scenarios in LLM medical evaluation.

### Closed-ended Tasks

Closed-ended medical questions are commonly used in medical education and assessment and play a crucial role in evaluating the performance of medical LLMs. Compared to open-ended questions, closed-ended questions typically have definite answers, reducing the possibility of subjective judgment. This enables easy quantification of the performance of LLMs on multiple-choice questions (MCQ), making them suitable for large-scale evaluation and comparison between different models. For instance, Royer et al.[12] developed an open-source evaluation toolkit called MultiMedEval to comprehensively assess the performance of LLMs on medical multiple-choice QA tasks. They conducted experiments on three datasets covering a wide range of medical domains, including general medical knowledge, radiological image interpretation, and yes/no questions from the literature. They compared the performance of closed-source and open-source models based on the BLEU scores. Liu et al.[38] evaluated models such as GPT-4 on the CMedExam dataset and found that its 61.6% accuracy was still significantly lower than human level (71.6%). The limitations mentioned above can be primarily attributed to the insufficient coverage of medical domain knowledge by LLMs, their limited understanding of professional medical terminology, and the inadequacies of current evaluation metrics. Li et al.[20] assessed ChatGPT's reliability and practicality in medical education by testing its performance on the 2021-2023 Chinese Master of Clinical Medicine comprehensive exams. The results found that both ChatGPT3.5 and GPT-4 passed the admission threshold but showed biases, with high accuracy in humanities subjects (93.75%) and lower accuracy in pathology (37.5% for ChatGPT3.5).

Another reason for using MCQs in evaluating LLMs is the broad knowledge coverage of existing MCQ

datasets. These datasets span a wide range of topics and difficulty levels, making them valuable resources for constructing domain-specific datasets. This is particularly advantageous when assessing the performance of language models in specialized fields, such as medicine, where comprehensive evaluation requires testing across diverse knowledge areas. Suchman et al.[39] evaluated the performance of GPT-3 and GPT-4 on the American College of Gastroenterology self-assessment tests. A score of 70% or higher is generally required to pass the evaluation. However, ChatGPT3.5 and ChatGPT4 scored 65.1 and 62.4, respectively, both failing to meet the passing threshold. Therefore, ChatGPT is currently not suitable for gastroenterology medical education in its present form. Bhayana et al.[40] evaluated ChatGPT's performance on radiology board-style exam questions without images. ChatGPT correctly answered 69% of the questions, essentially passing the exam. Gupta et al.[41] evaluated ChatGPT's performance on the Plastic Surgery Inservice Training Examination to determine if it could serve as an assistive tool for resident education. The model achieved an accuracy of 54.95% on this task. By testing ChatGPT's performance on various questions related to general knowledge, information clarification, case-based learning, and evidence-based medicine promotion, the results suggest that ChatGPT could potentially become an effective tool for plastic surgery resident education.

Despite closed-ended questions providing an objective and quantifiable means to evaluate LLMs' medical knowledge, there are limitations, including a focus on procedural knowledge, a lack of deep assessment of complex situations, and a failure to reflect the models' performance in real-world scenarios. Li et al.[42] found that LLMs such as GPT and InstructGPT showed sensitivity to answer positions in bilingual MCQs, especially a tendency to choose answers located in the first position, revealing potential biases. The findings suggest that current MCQ benchmarks may not accurately reflect the true capabilities of LLMs, indicating a need for more robust assessment mechanisms to measure LLM performance. Therefore, a comprehensive evaluation framework should include both closed-ended and open-ended tasks to assess the full capabilities and limitations of medical LLMs.

### Open-ended Tasks

Open-ended tasks refer to generating answers in a diverse way. Compared to the multiple-choice tasks have only a few simple and mechanical answers like “yes or no”, open-ended tasks require improved natural language processing ability and feasibility of LLMs to fully express opinions. For this reason, the evaluation dimensions become more complicated in open-ended tasks accordingly. In medical fields, there are three main open-ended tasks for LLMs: summarization, information extraction and medical QA.

***Summarization***

Text summarization related to the medical field requires extracting key information from various medical data sources such as medical literature and electronic health records (EHRs) and then generating a short concise summary of the given medical text. LLMs can help summarize medical research evidence.[43,44] Tang et al.[45] investigated the capabilities and limitations of LLMs in summarizing reviews across six clinical domains. Automatic metrics were used to assess the lexical and semantic similarity and human evaluation was conducted on coherence, factual consistency, comprehensiveness, and harmfulness. The finding revealed that LLMs could be susceptible to generating incorrect information and lead to potential harm due to misinformation. Hake et al.[46] evaluated ChatGPT's ability to summarize abstracts from more types of clinical research such as case series, observational studies, randomized controlled trials, and more. They solely relied on human assessment, encompassing quality, accuracy, bias, and relevance and showed LLMs' acceptable performance in these aspects.

As for summarizing EHRs, LLMs have the potential to be applied in generating medical text to ease the documentation burden for physicians.[47-49] Dubinski et al.[50] investigated the time consumption and factual correctness of neurosurgical discharge summaries and operative reports ChatGPT generated. The result showed significant time reduction and a high degree of factual correctness with the assistance of ChatGPT. Zaretsky et al.[51] further assessed the ability of LLMs to transform discharge summaries into patient-friendly language and format from readability, accuracy, and completeness. These findings suggested that despite not being perfect and containing a few inappropriate omissions or insertions, LLMs have the potential to enhance the efficiency of generating medical documents. In addition to generating discharge summaries and various other documents, EHRs play a crucial role in medical examination reports, which often involve complex terminology. LLMs can summarize the key information and convey the message in plain language. Lyu et al.[52] and Chung et al.[53] evaluated ChatGPT's performance in summarizing radiology reports including CT and MRI. Lyu et al.[52] assessed completeness and correctness while Chung et al.[53] assessed readability, factual correctness, ease of understanding, completeness, potential for harm and overall quality. They demonstrate the novel feasibility of using LLMs to generate patient-friendly summaries of radiology reports. Van et al.[54] evaluated LLMs' performance in clinical text summarization across multiple tasks including radiology reports, patient questions, progress notes, and doctor-patient dialogues from similarity, completeness, correctness and conciseness. They provided evidence of LLMs outperforming medical experts in clinical text summarization across multiple tasks.

Notably, the majority of studies in their conclusion sections mentioned potential ethical issues and safety concerns that may arise in clinical applications, which constitute an integral aspect not to be overlooked in future clinical implementations.

### *Information extraction*

Information extraction is basic but essential in medical fields, especially in biomedical domain. Numerous studies showed LLMs have the potential to assist researchers or patients search and acquire knowledge from a large amount of biomedical data.

Named entity recognition (NER) is one of the information extraction tasks that involves identifying named entities such as genes, proteins, diseases, and more, in the given input text. Many studies evaluated LLMs' performances on this task in the first place.[55,56] Evaluation metrics such as precision, recall, and F1 score were widely used in the evaluation of this task. In the EHR field, Gu et al.[57] trained an LLM to extract and quantify stroke severity from EHR based on Chinese clinical named entity recognition. This model demonstrated a high F1 score of 0.990 which ensured the reliability of the model in accurately extracting the entities for the subsequent automatic NIHSS scoring. In addition, Guevara et al.[58] used LLMs to extract social determinants of health (SDoH) categories from real-world clinic notes. Their fine-tuned models achieved higher F1 scores than ChatGPT for most SDoH classes and highlighted the potential of LLMs to improve real-world data collection and identification of SDoH from the EHR.

Relation extraction refers to extracting relations between named entities in a given text such as relations between genes and diseases, or genes and proteins. Researchers also evaluate the performance of LLMs using statistical measures such as accuracy, precision, recall, and F1 scores. Most researchers evaluate LLMs on relation extraction through datasets related to drug-drug interaction, chemical-disease relation, gene-disease associations[55] and drug-target interaction[56,59]. Cinquin et al[60] reported ChIP-GPT which was able to extract data from biomedical database records especially to identify cell lines, the ChIP target and the gene corresponding to the target. ChIP-GPT demonstrates 90-94% accuracy when trained with 100 examples.

***Medical question answering***

Patients often have numerous questions and concerns related to health and medical clinical information. A number of studies evaluated LLMs' open-minded responses across various disciplines, encompassing questions such as the concept of disease, etiology, examination, diagnosis, prevention, treatment and care. Majority of LLMs evaluations concentrated on clinical queries for the reliability of the responses or advises LLMs provided which may impact their healthcare decision to a great extent.

The evaluation of LLMs in open-ended tasks involves diverse sources of questions, including author-designed questions,[61-63] questions from professional societies and institutions,[64,65] questions from social media,[66,67] and questions from validated real or simulated clinical patient cases[68-70]. When it comes to specific applications in various clinical disciplines, clinicians paid more attention to the LLMs application in their clinical sub-specialty. Take ophthalmology for example, Ali[71] evaluated ChatGPT's performance on queries related to lacrimal drainage disorders from correctness, but the performance of ChatGPT in the context of lacrimal drainage disorders, at best, can be considered average. The study highlighted there is a need for it to be specifically trained for individual medical subspecialties. Potapenko et al.[72] consult ChatGPT common retinal diseases including age- related macular degeneration, diabetic retinopathy, retinal vein occlusion, retinal artery occlusion and central serous chorioretinopathy and evaluated the accuracy of responses. ChatGPT provided highly accurate responses in most questions except for questions dealing with treatment options. Rasmussen et al.[73] evaluated the accuracy of responses to typical patient-related questions on vernal keratoconjunctivitis. The result also demonstrated that responses to treatment/prevention questions obtained lower scores than the rest. We summarize many evaluations of LLMs' performance in answering medical questions through multiple medical disciplines (**Table 1**). Notably, the types of questions are similar between studies: general summary, etiology, prognosis, treatment, and prevention which is a series of cross-sectional questions patients want to know.

The dimensions used to assess performance in medical QA tasks vary across the literature. However, relying solely on metrics like BLEU, CIDEr, ROUGE, and others for automatic evaluation of these tasks has its limitations. Interestingly, a few studies have employed GPT-4 to automatically evaluate responses of other models across multiple dimensions.[74] This introduced a novel and user-friendly automated assessment method. However, further exploration is needed to determine the reliability and consistency of GPT-4 compared to human assessments. Currently, human evaluation is necessary due to the inability of parsing scripts to properly identify the correctness of responses with the latest updates in medical knowledge and analyze the implications of these responses for patients. In addition to the crucial evaluation metric of accuracy or correctness, which is prioritized in most studies, there are other valuable evaluation metrics such as completeness[63,64,66,67,75-80] and readability[63,75,76,81-83]. Some studies have also examined relatively uncommon dimensions, including safety and humanistic care. Cadamuro et al.[84] evaluated whether the responses were safe which referred to the potential negative consequences and detrimental effects of ChatGPT's response on the patient's health and well-being. Menz et al[85] explored the issue of the security of LLMs when LLMs were prompted to generate health disinformation and measured whether safeguards of LLMs prevented the generation of health disinformation. This study found that inconsistencies in the effectiveness of LLM safeguards to prevent the mass generation of health disinformation. Yeo et al.[67] and Zhu et al.[63] paid attention to LLMs' performance in emotional support (or humanistic care) which referred to offering psychological assistance to individuals facing particular emotional distress when diagnosed with cancer or other diseases. Their results showed that ChatGPT demonstrates empathy when answering patients' questions.

**Image Processing Tasks**

Medical diagnosis and treatment often rely on various types of images, including CT and MRI in radiology, as well as fundus photographs and OCT in ophthalmology. By jointly modeling image and text information, LLMs can not only better understand the content of medical images but also generate diagnostic reports based on images. This section will systematically review the evaluation progress of LLMs in medical image processing, focusing on tasks including image classification, report generation and visual question answering (VQA). Examples of various image processing tasks can be found in **Figure 3**.

### *Image Classification*

Medical image classification is a fundamental task for evaluating the ability of LLMs to understand medical image content and identify disease patterns. This task demands strong feature extraction capabilities from the model to distinguish subtle differences between images. Common performance metrics include accuracy, recall, precision, and F1 score. Royer et al.[12] evaluated the multi-class/multi-label classification performance of LLMs using metrics such as F1, AUROC, and Accuracy on 15 datasets including MIMIC-CXR, Pad-UFES-20, and CBIS-DDSM. The results showed that LLMs performed well on different medical image tasks but still need improvement in recognizing fine-grained visual concepts. Yan et al.[86] proposed a new paradigm for medical image classification that integrates natural language concepts. Evaluations on multiple datasets show that this method effectively reduces false correlations and enhances the robustness and interpretability of the model. However, it has limitations such as reliance on potentially biased outputs from LLMs, high dependence on the accuracy of pre-trained models, and the need for multiple rounds of revisions and prompts. Wan et al.[87] proposed the Med-UniC framework, which employs cross-lingual text regularization techniques. This framework not only exhibits outstanding performance on multiple medical image classification datasets, such as CheXpert, RSNA, and COVIDx but also achieves remarkable results in zero-shot classification tasks. In the cross-lingual evaluation on the CXP500 and PDC datasets, Med-UniC, utilizing both English and Spanish prompts, achieved an improvement of over 20% in F1 scores, demonstrating its effectiveness and adaptability when processing non-English community images and prompts. These results underscore the importance of reducing community bias to enhance the diagnostic quality and task performance of the model in clinical applications. Moreover, these findings further highlight the robust feature extraction and cross-lingual understanding capabilities of vision-language models.

In summary, LLMs have shown promising results in image classification tasks, especially when incorporating language information, and their performance has been significantly improved. Future Research could focus on how to effectively align language prompts and images, as well as zero-shot generalization capability.

### *Report Generation*

Medical report generation aims to automate the process of generating diagnostic reports based on medical images. Given medical images, LLMs are required to generate diagnostic reports that conform to clinical norms, including major sections such as imaging findings, diagnostic opinions, and differential diagnoses. It requires not only accurately extracting key findings from images but also standardized descriptions using professional terminology and reasoning about possible diagnoses.

Currently, the evaluation on this task is mainly carried out on large-scale radiology datasets such as IU X-Ray, MIMIC-CXR, and CX-CHR. A series of language metrics such as BLEU, ROUGE, METEOR and CIDEr are used to comprehensively examine the fluency, relevance and readability of the generated reports. Li et al.[88] evaluated the radiology report generation performance of GPT-4V. The results showed that the model could generate descriptive reports with structured prompts, but there is still space for improvement in the accuracy of generating specific medical terms. Royer et al.[12] systematically tested and demonstrated the ability of LLMs to generate diagnostic reports based on chest X-rays using the MIMIC-CXR dataset. Wang et al.[89] proposed the R2GenGPT model and validated its performance in radiology report generation. Lee et al.[90] constructed the LLM-CXR model and demonstrated its ability to accurately capture lesion characteristics, locations, and severity, closely aligning with the input text. Both methods are promising solutions for automating and improving radiology reports, but no human evaluation has been conducted. At the same time, Some researchers have conducted comprehensive evaluations of human-computer interaction. Chen et al.[91] introduced the ICGA-GPT model and assessed its ability to generate ophthalmic reports through automated and manual evaluations. The results showed that ICGA-GPT achieved satisfactory scores in BLEU1-4, CIDEr, ROUGE-L, and SPICE metrics. Furthermore, subjective assessments conducted by ophthalmologists indicated high accuracy and completeness scores for the model.

Although the medical report generation task has shown progress, it still faces many challenges. First, the reports generated by existing models are still difficult to fully meet clinical standards in terms of professionalism and accuracy, lacking systematic human evaluation and verification. Second, the interpretability of the report generation process is insufficient, making it difficult for doctors and patients to trust it. In addition to technical metrics, human evaluation by medical experts is necessary to assess the factual correctness and clinical usability of the generated reports.

### *Visual Question Answering*

Medical VQA is a key task for evaluating the multimodal integration and reasoning capabilities of LLMs. In this task, given a medical image and related questions, LLMs need to comprehend the contextual information of the questions and generate answers that conform to clinical facts and logic. In contrast to image classification and report generation, VQA requires stronger cross-modal understanding capabilities and higher demands on natural language expression. Additionally, the open-ended nature of VQA questions demands that the model possess higher levels of creativity and flexibility, encompassing both image comprehension and extensive medical knowledge beyond the image. This puts forward high requirements for the model in terms of visual understanding, language analysis, knowledge reasoning, and other aspects, and is an important standard for examining its understanding ability in medical scenarios.

Due to the difficulty of medical VQA tasks, constructing large-scale, high-quality datasets is crucial for evaluating model performance. Lin et al.[92] systematically reviewed representative datasets in this field. such as VQA-Med 2018, VQA-RAD, VQA-Med 2019, RadVisDial, PathVQA, VQA-Med 2020, SLAKE, and VQA-Med 2021. These datasets cover various medical domains, including radiology and pathology, providing a foundation for comprehensively evaluating the multimodal analysis capabilities of LLMs. Li et al.[88] evaluated the capability of GPT-4V in medical VQA tasks. The results indicated that the model excelled in distinguishing question types but did not meet existing benchmarks in accuracy. The advantage lies in its strong language understanding and question classification abilities, while the limitation is the underutilization of medical image information. However, some medical specialties, such as ophthalmology, still lack large-scale VQA datasets. Mihalache et al.[93] evaluated the performance of the artificial intelligence chatbot ChatGPT-4 in interpreting a dataset of multimodal ophthalmic images. The study results demonstrated that ChatGPT-4 performed well in the task of identifying types of ophthalmic examinations, achieving an accuracy rate of 70%. However, when it came to lesion identification, the model's average accuracy was only 65%, indicating the limitations of ChatGPT-4 in accurately recognizing and describing lesions from ophthalmic images. Xu et al.[94] tested the ability of the GPT-4V model on the VQA task on datasets of multiple ophthalmic imaging modalities. The datasets included slit-lamp, scanning laser ophthalmoscopy, fundus photography, optical coherence tomography, fundus fluorescein angiography, and ocular ultrasonography images. The results showed that GPT-4V performed well in the task of identifying ophthalmic examination types, with an accuracy of 95.6%. However, in terms of lesion identification, the model's average accuracy was only 25.6%, indicating limitations of GPT-4V in accurately identifying and describing lesions from ophthalmic images. Therefore, constructing high-quality VQA datasets within each medical specialty is of great significance for comprehensively evaluating the performance of LLMs.

Evaluation of VQA tasks currently employs a comprehensive approach by combining both automated and manual assessments to thoroughly examine the quality of generated answers. Automated evaluation includes classification metrics such as accuracy, F1 score, and language metrics like BLEU, enabling large-scale assessment of answer accuracy and natural language generation capabilities. On the other hand, manual evaluation allows for a more detailed and personalized inspection of response expertise and appropriateness. Royer et al.[12] adopted multi-angle evaluation metrics including precision, recall, BLEU, etc. on datasets such as Path-VQA, SLAKE, and VQA-Rad, and deeply analyzed the processing capabilities of LLMs on different types of questions. This evaluation system helps to clarify the advantages and limitations of current technologies and indicates the direction for subsequent algorithm improvements. Wu et al.[95] conducted tests on LLMs and discovered that these models are at risk of generating hallucinations in Medical VQA, meaning they may produce coherent but factually incorrect responses. They created a hallucination benchmark consisting of medical image and question-answer set pairs and performed a comprehensive evaluation, revealing the limitations of current models and analyzing the effectiveness of various prompting strategies. The research emphasizes the importance of reducing hallucinations in medical VQA systems to improve accuracy and safety.

Overall, although there has been some progress in evaluating LLMs in medical VQA tasks, they still face several challenges. Firstly, the lack of standardized evaluation datasets makes it difficult to compare and assess performance. Different research teams may utilize different datasets or create their own, thereby limiting the comparability of results. Secondly, in the medical field, the interpretability and credibility of the models are of utmost importance. Evaluating the performance of LLMs in medical VQA tasks necessitates considering the interpretability and confidence of the answers they provide, ensuring that the models can deliver explanations for their answers and express their uncertainty.

***Others***

In addition to the application in scenarios such as image classification, report generation and VQA, LLMs also show good prospects in other medical image analysis tasks such as medical image segmentation and cross-modal retrieval. In medical image segmentation, LLMs can assist by interpreting key information from medical reports or prompts and integrating this knowledge to guide image segmentation algorithms, enhancing their accuracy and efficiency. For cross-modal retrieval tasks, LLMs act as a bridge by translating natural language commands into actionable signals for image processing models. Bai et al.[96] provide early examples of such evaluations, selecting 2000 image-text pairs with four difficulty levels from M3D-Cap for text-to-image retrieval (TR) and image-to-text retrieval (IR) tasks. Evaluation metrics include recall at ranks 1, 5, and 10 for IR and TR, assessing the model's ability to retrieve relevant images or texts from top-ranked results. For segmentation tasks, the M3D-Seg dataset combines data from various sources such as AbdomenCT-1K, Totalsegmneter, and CT Organ, with Dice used as the metric. Current evaluations of LLMs in these new tasks are inadequate, which shows a necessity for a deeper investigation into evaluation benchmarks. Future research should enhance LLM assessment by employing varied datasets and detail metrics, and by focusing on improving model interpretability and explainability to facilitate clear decision-making.

In summary, LLMs have made encouraging progress in multiple tasks of medical image processing, demonstrating abilities from image recognition to semantic understanding and cross-modal reasoning. However, the current evaluation work is still mainly focused on performance assessment, with insufficient examination of model interpretability, privacy security and ethical issues. Future research should not only further expand application scenarios and optimize model performance but also pay attention to humanistic care, strengthen human-computer collaboration, and promote the credible and beneficial application of large models in the medical field.

**Real-world Medical Scenarios**

In most LLMs' evaluation scenarios, although real-world problems and data from clinical practice are introduced, the complexity of the conversation or data in real clinical settings far exceeds that which can be compared with organized datasets. Moreover, in real clinical settings, patients and doctors are the two most important roles, yet there is currently a significant lack of assessment for true human-machine interaction (such as between doctors and LLMs, or patients and LLMs). How do doctors or patients directly perceive the use of LLMs? How do LLMs perform after direct interaction with patients or doctors? This is an indispensable aspect for the future application of LLMs in clinical practice.

Bao et al.[97] used both single-turn and multi-turn consultation to simulate real-world consultation scenarios. They systematically assessed DISC-MedLLM they proposed from four metrics: proactivity, accuracy, helpfulness, and linguistic quality and employed GPT-3.5 to play the role of the patient and chat with the model for three rounds interestingly.  To explore patient perspectives, Mannhardt et al.[98] recruited 200 participants and randomly assigned three clinical notes about breast cancer with varying levels of modifications using LLM. Participants evaluated the level of understanding and acceptance of clinical notes. In-depth interviews of seven self-identifying breast cancer patients were also conducted via video conferencing to assess their trust in clinical notes modified by LLM. Additionally, Tu et al.[99] conducted a randomized crossover study of blinded consultations in the style of a remote OSCE to evaluate LLMs' performance. Objective Structured Clinical Examination (OSCE) is an objective and organized assessment framework which provides a simulation of real-world clinical scenarios to assess the clinical competency of medical students Patient actors consulted with real physicians and LLMs in random order and the conversations were assessed by patient actors and specialists respectively.

There is still a lack of real-world medical scenario evaluation cases due to concerns about privacy data.

Given the increasingly enhanced capabilities of LLMs, Researchers need to shift benchmarks from static datasets to dynamic simulation environments or real-world scenarios.[100] It's necessary to conduct randomized controlled trials (RCTs) to compare whether LLM truly aids patients and assists in clinical practices in the future.

## EVALUATION METHOD

When evaluating LLMs in the field of medicine, it is necessary to consider both the performance of the model and its potential impact on patient health. This process involves not only automated assessments to quantify the model's task-specific capabilities but also manual evaluations to measure the quality, accuracy, and applicability of the model's outputs to real medical scenarios. The following will provide a detailed explanation of these two evaluation methods.

### Automatic Evaluation

Automatic evaluation focuses on objectively assessing the performance of LLMs through automatic algorithms. In classification tasks, metrics such as accuracy, specificity, precision, sensitivity and F1 score are used to quantify the performance of model predictions.[101] Duong et al. compared the accuracy of ChatGPT and human respondents in answering genetic questions.[28] Cai et al. evaluated Bing Chat, ChatGPT 3.5, and 4.0 in answering 250 questions in basic and clinical science self-assessment projects, with the primary outcome being response accuracy.[22]

For long-text tasks, several metrics are used to evaluate the quality of generated text. Metrics such as BLEU, ROUGE, CIDEr, and METEOR focus on text overlap and assess the literal accuracy of the generated text.[26,91,102] On the other hand, metrics like BERTScore and MoverScore measure semantic similarity, evaluating the semantic accuracy and consistency of expression.[103-105] In the context of evaluating medical text generation tasks in open-ended QA scenarios, specialized tools like QAEval and QAFactEval have also been employed.[44] To assess the fluency and readability of the text, several commonly used metrics are available, including the Flesch Reading Ease (FRE) Score, Flesch-Kincaid grade level (FKGL), Gunning Fog Index (GFI), Coleman-Liau Index (CLI), and Simple Measure of Gobbledygook (SMOG).[35] These metrics are utilized to determine whether the medical content generated by the language model is informative, effective in conveying information, and user-friendly.

**Human Evaluation**

Automated evaluation methods fall short in covering all essential aspects, particularly in sensitive domains like medicine that require advanced knowledge and ethical judgment, making manual evaluation crucial in such cases.[26] One relatively simple way to facilitate manual evaluation is to use qualitative methods such as case studies.[106] These allow manual evaluators to carefully compare the content of the LLM with the ground truth, thereby revealing subtle differences that automated evaluation methods cannot identify.

To facilitate scientific review and statistical analysis of LLM outputs by manual evaluators, various scoring protocols have been adopted to assess the quality of generated response. Standardized scales are applied, including the DISCERN scale,[107] the JAMA benchmark criteria,[108] the Global Quality scale,[109] and others. the Likert scale, and custom grading criteria. Among these, the Likert scale is the most commonly used. Another approach is to identify and statistically analyze the occurrence probabilities of different types of predefined errors, such as factual errors and logical errors.[102,110] To diversify evaluation modes, some studies use custom grading rules or Likert-style rules to assess text quality across multiple levels. The Likert scale is widely used in social science and psychological research to evaluate people's views or attitudes toward specific viewpoints.[111] In the field of medicine, each evaluation dimension to be considered can be transformed into a series of statements, and corresponding answer options can be provided to investigate the degree of identification of respondents with different dimensions of performance. For example, Samaan et al.[29] recruited board-certified bariatric surgeons and used a 4-point scale to evaluate both the accuracy and comprehensiveness of ChatGPT, where 1 represented comprehensive, 2 represented correct but insufficient, 3 represented partially correct and partially incorrect, and 4 represented completely incorrect. Chen et al.[91] used a 5-point Likert scale ranging from 1 (strongly disagree) to 5 (strongly agree) to evaluate the completeness and correctness of reports generated by the model.

Additionally, some studies explore more comprehensive evaluation schemes by incorporating expanded dimensions, adversarial evaluation, and side-by-side comparisons to achieve finer-grained evaluation results. Singhal et al.[4] created a comprehensive evaluation framework involving 12 aspects, including scientific consensus, extent of possible harm, likelihood of possible harm, evidence of correct, comprehension, evidence of correct retrieval, evidence of correct reasoning, evidence of incorrect comprehension, evidence of incorrect retrieval, evidence of incorrect reasoning, inappropriate/incorrect content, missing content, possibility of bias. This framework serves as an important reference for multidimensional assessment. The team later also designed adversarial datasets to evaluate their improved LLMs, which helps identify model weaknesses when faced with challenging inputs deliberately constructed to be misleading, such as rare diseases or setting traps involving medical safety, to enhance the model.[112] In side-by-side comparison, human reviewers are asked to simultaneously compare the results generated by two or more models and rank them according to predefined criteria.[113] This method eliminates the influence of personal preferences or subjective differences among the reviewers when compared with direct scoring. Tu et al.[110] used this method to systematically compare models with different scaling and accurately determine their relative strengths and weaknesses.

Lastly, collecting opinions from evaluators at various levels can enhance the comprehensiveness of the evaluation process. Currently, the assessment of most studies is conducted by professional physicians.[52,114,115] However, relying exclusively on doctors as evaluators may not align with the development of patient-centered medical LLMs. Therefore, some studies have also included non-professionals, such as patients and the general public, to participate in evaluating the LLMs.[30,33] For instance, Singhal et al. engaged five non-medical Indian raters to analyze the usefulness and practicality of LLM's responses to long-form questions.[4] This approach enables the capture of details that may be overlooked when solely relying on expert perspectives.

## LIMITATION AND FUTURE DIRECTION

To ensure the reliable application of LLMs in the medical field, constructing precise and effective evaluation frameworks is crucial. This section delves into the challenges currently faced by the evaluation of medical LLMs and proposes strategies for future development (**Figure 4**).

Firstly, establishing suitable evaluation frameworks requires the preparation of appropriate datasets. Although datasets for categorization and report generation are relatively abundant, there is a lack of high-quality medical QA datasets. In particular, crafting high-quality VQA datasets poses a significant challenge; they require the selection of medically compliant images, including diverse types of questions and answers. Furthermore, the construction of evaluation datasets should not only encompass a rich understanding of professional medical knowledge but also take into consideration medical safety and ethical issues. This process demands not only data selection from real-world clinics but also a substantial involvement of medical professionals. However, given that medical professionals have limited time and energy, primarily focused on clinical work and research, the additional burden of participating in dataset construction is often impractical.[116] Therefore, future studies may explore the combination of human and LLM capabilities to collaboratively build and continuously optimize evaluation datasets, thereby enhancing the scalability of datasets and reducing the manual workload.

Secondly, setting comprehensive and effective assessment standards is also a key component in the current evaluation frameworks. Evaluation dimensions traditionally concentrate on accuracy and completeness, which are indeed crucial in the medical field as they relate to the model's ability to prevent misdiagnoses and missed diagnoses. However, performance aspects such as safety, empathy, and interpretability must also be considered now more than ever.[117-119] Regarding assessment metrics, the focus remains predominantly on traditional classification metrics and NLP metrics. Some studies have attempted to utilize advanced LLMs like GPT4 for automated assessment.[120-122] Yan et al. have proposed an automatic evaluation model based on MedLLaMA, which aims to assess the correctness, expertise, and completeness of answers in open clinical scenarios.[123] However, the stability and robustness of these automatic evaluation models have not been confirmed yet. Appropriate response is not limited to correctness but also considers applicability in different contexts. Responses that deviate from the standard answer might be appropriate, while responses that are similar or consistent with the ground truth could still be misleading. Therefore, in the short term, we cannot entirely dispense with human evaluation. The future direction should involve further development and validation of automated evaluation systems—that balance safeguarding rational assessments while alleviating pressure on human resources and facilitating large-scale evaluations.

As for the practical application of LLMs, most studies are currently at the preclinical validation stage.

Actively promoting rigorous clinical trials to verify the actual utility of LLMs in medical applications is extremely critical. Trial designs need to conform to real-world clinic and compare LLMs to existing practices—including other health care systems, traditional AI tools, and healthcare professionals of different levels—to truly assess their value in practical applications. Appropriate endpoints, such as reducing morbidity, improving work efficiency, and patient or physician satisfaction, are required to gauge success or failure. The design of LLM interventions in clinical trials can benefit from the application of non-LLM chatbots in randomized controlled trials.[5] When designing an assessment framework for medical LLMs, it is crucial to ensure diversity and personalization in the selection of evaluators. This includes not only physicians but also incorporating perspectives from patients, medical students, and other real users, based on specific application scenarios and functionalities. In fact, generalist LLMs have experimented with collecting user feedback for online assessments.[124] They analyze service logs and directly or indirectly assess user satisfaction, enabling close-to-real-world scenario efficacy assessment. This method not only garners valuable and actual user feedback but also provides an ideal path for continuous performance monitoring, which could also be expected to apply to LLMs in the medical field.

In summary, a precise and effective evaluation framework is indispensable for the assessment of medical LLMs. Recently, Abbasian et al. proposed a five-element evaluation framework including models, environments, interfaces, interactive users, and leaderboards,[125] offering valuable references for research and applications in the sector.

## CONCLUSION

We provide an overview of the recent advancements in evaluating LLMs in the medical field, with a special focus on key elements of the evaluation framework, including datasets, evaluation methods, dimensions, and scenarios. Future research should harness the interdisciplinary expertise of medical professionals and computer scientists to address the existing challenges in various domains, and ultimately optimize the application of LLMs to enhance the quality and efficiency of medical services and patient experience.

## COMPETING INTERESTS

There are no conflicts of interest to declare by the authors.

## FUNDING/SUPPORT

Danli Shi and Mingguang He disclose support for the research and publication of this work from the Start-up Fund for RAPs under the Strategic Hiring Scheme (Grant Number: P0048623) and the Global STEM Professorship Scheme (Grant Number: P0046113) from HKSAR. The sponsor or funding organization had no role in the design or conduct of this research.

## FIGURE LEGENDS AND TABLES

**Figure 1.** Illustration of the potential LLM evaluation framework in medicine.

**Figure 2.** Summary of data sources and evaluation dimensions related to different task scenarios in LLM medical evaluation.

**Figure 3.** Examples of various image processing tasks.

**Figure 4.** Overview of challenges and future directions of LLM evaluation in medicine.

**Table 1.** Major studies on the evaluation of LLMs in medical question answering tasks.

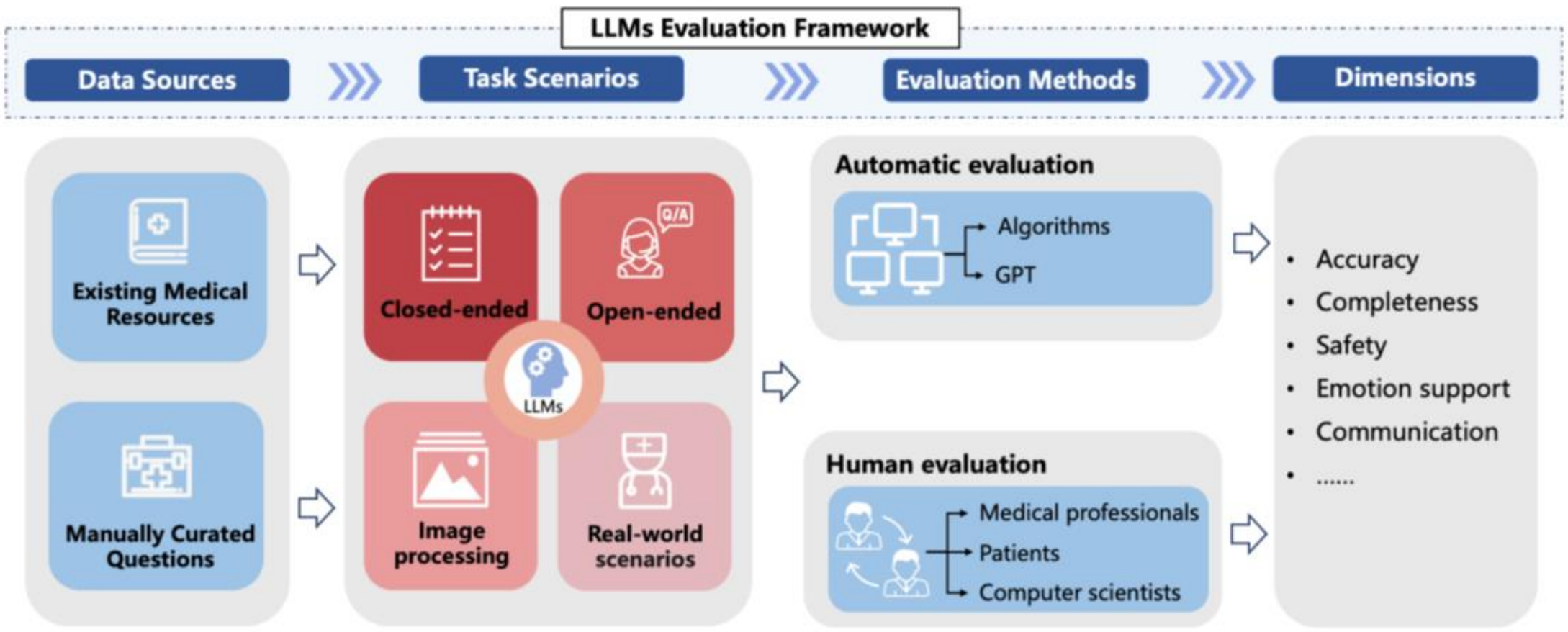

**Figure 1**. Illustration of the potential LLM evaluation framework in medicine.

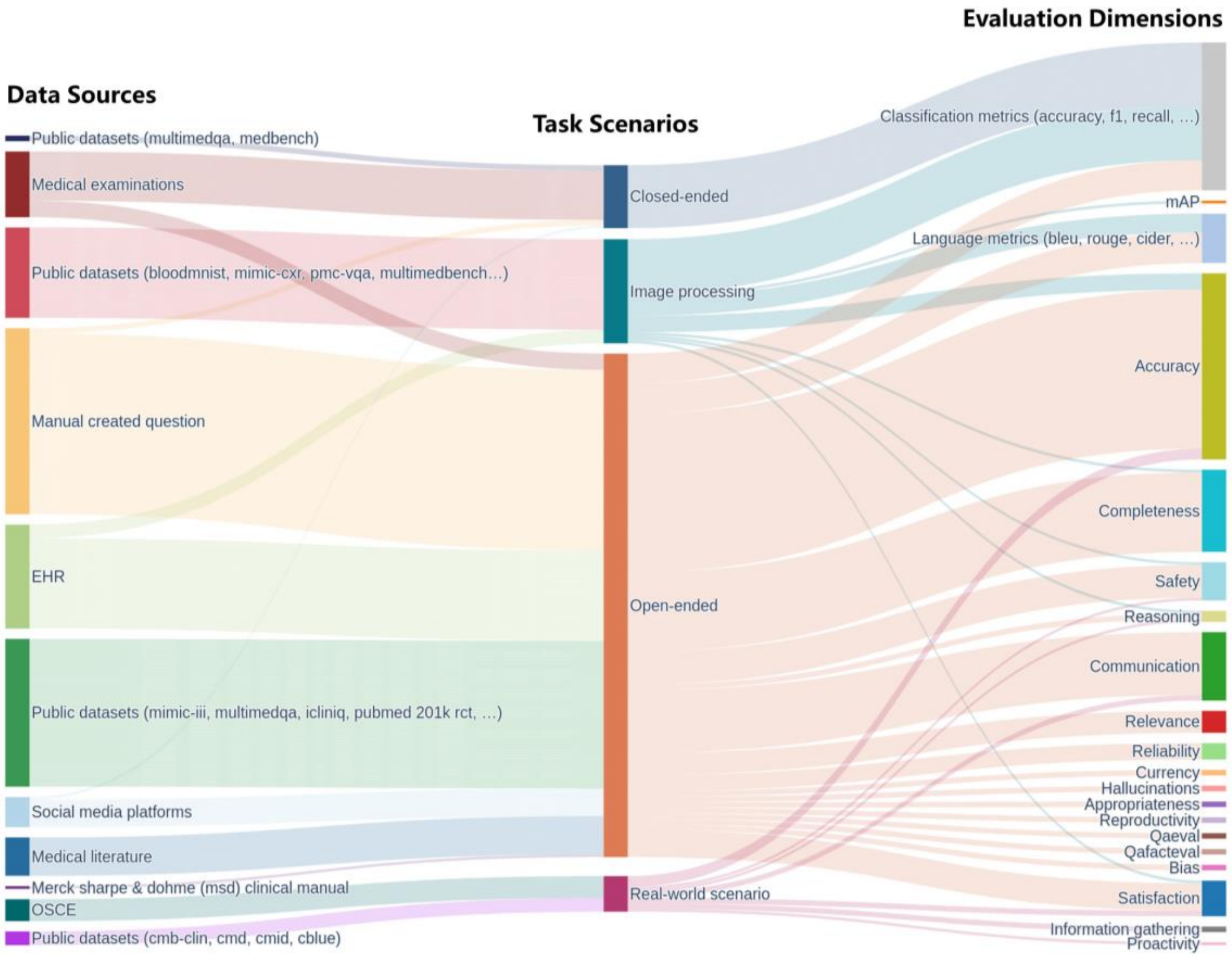

**Figure 2**. Summary of data sources and evaluation dimensions related to different task scenarios in LLM medical evaluation.

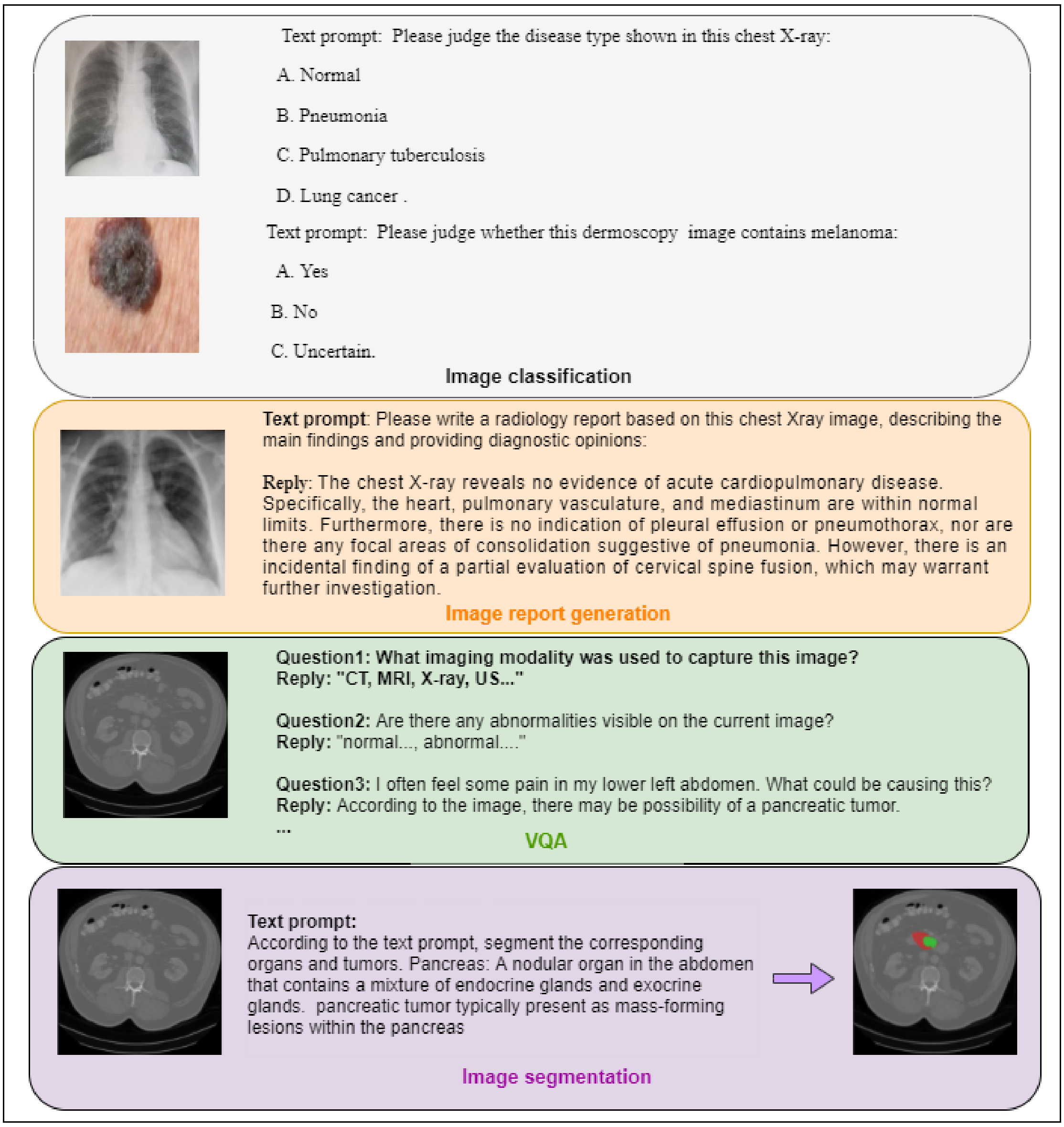

**Figure 3.** Examples of various image processing tasks.

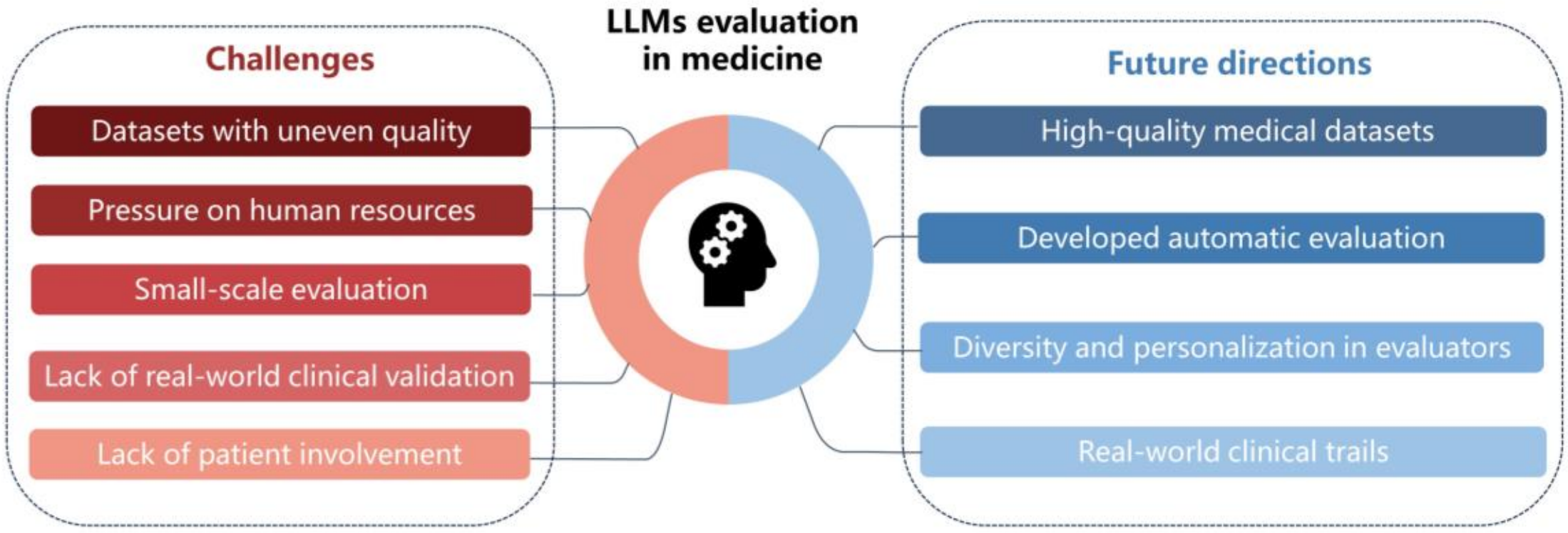

**Figure 4.** Overview of challenges and future directions of LLM evaluation in medicine.

**Table 1.** Major studies on the evaluation of LLMs in medical question answering tasks.

| study | Question domain | Question source | Study design | Evaluation method | accuracy | completeness | safety | readability | reproducibility | Other dimension | Inter-rater consistency |
|---|---|---|---|---|---|---|---|---|---|---|---|
| **Internal Medicine** | | | | | | | | | | | |
| **Sarraju et al[61]** | Cardiovascular Disease | Author-designed | Non-comparative study | Direct evaluation without scale | | | | | √ | appropriate | |
| **Johnson et al[81]** | Oncology | Professional institutions | Non-comparative study | Direct evaluation without scale | √ | | | √ | | Word count | √ |
| **Howard et al[82]** | Infectious Diseases | Author-designed | Non-comparative study | Direct evaluation without scale | √ | | | √ | √ | | |
| **Ayers et al[66]** | General Internal Medicine | Social media | Cross-sectional comparative study | Self-designed Likert scale, evaluator preference | √ | √ | | | | Quality, word count, humanistic care | |
| **Lee et al[76]** | Gastroenterology | Professional institutions | Non-comparative study | Self-designed Likert scale | √ | √ | | √ | | Satisfaction | |
| **Sarink et al[126]** | infectious diseases | Clinical patient cases | Non-comparative study | Self-designed Likert scale | √ | | | | | Reference generation | √ |
| **Onder et al[127]** | Endocrinology | Professional society | Non-comparative study | Standardized scale | √ | | | √ | | Quality | |
| **Ge et al[62]** | Hepatology | Author-designed | Cross-sectional | Self-designed Likert scale | √ | √ | √ | | | | |
| | | | comparative study | | | | | | | | |
| **Civettini et al**[69] | Hematology | Clinical patient cases | Cross-sectional comparative study | Direct evaluation without scale | √ | | | | | | |
| **Surgery** | | | | | | | | | | | |
| **Zhu et al**[63] | Urology | Author-designed | Cross-sectional comparative study | Self-designed Likert scale | √ | √ | | √ | √ | Humanistic care | |
| **Haver et al**[128] | Breast Surgery | Author-designed | Non-comparative study | Direct evaluation without scale | | | | | √ | Appropriate | |
| **Sorin et al**[70] | Breast Surgery | Clinical patient cases | Cross-sectional comparative study | Self-designed Likert scale | | | | | | agreement | √ |
| **Yeo et al**[67] | Hepatic disease | Socia media and professional societies and institutions | Non-comparative study | Self-designed Likert scale | √ | √ | | | √ | Emotion support | |
| **Cao et al**[78] | Hepatic disease | Author design | Non-comparative study | Self-designed Likert scale | √ | √ | | | | | |
| **Xie et al**[64] | Plastic Surgeon | Professional society | Non-comparative study | Direct evaluation without scale | √ | √ | | | | | |
| **Hurley et al**[129] | Arthropathy | Author-designed | Non-comparative study | Standardized scale | √ | | | √ | | Quality | |
| **Other disciplines** | | | | | | | | | | | |
| **Cadamuro et al**[130] | laboratory medicine test | Author-designed | Non-comparative study | Self-designed Likert scale | √ | | √ | | | Relevance, helpfulness | |
| **Young et al**[83] | Dermatology | Author design and professional society | Non-comparative study | Self-designed Likert scale | √ | | | √ | | | √ |
| **Ali et al**[71] | Ophthalmology | Author-designed | Non-comparative study | Direct evaluation without scale | √ | | | | √ | | |
| **Potapenko et al**[72] | Ophthalmology | Author-designed | Non-comparative study | Self-designed Likert scale | √ | | | | | | √ |
| **Marshall et al**[131] | Ophthalmology | Author-designed | Non-comparative study | Self-designed Likert scale | √ | √ | | | | | √ |
| **Bernstein et al**[132] | Ophthalmology | Social media | Cross-sectional comparative study | Self-designed Likert scale, evaluator preference | √ | | √ | | | | |
| **Caranfa et al**[133] | Ophthalmology | Social media | Non-comparative study | Direct evaluation without scale | √ | | | | √ | | |

| | | | | | | | | | | |
|---|---|---|---|---|---|---|---|---|---|---|
| **Momenaei et al[134]** | Ophthalmology | Author-designed | Non-comparative study | Direct evaluation without scale | | | | √ | | Appropriateness |
| **Huang et al[65]** | Ophthalmology | Professional society and clinical patient cases | Cross-sectional comparative study | Self-designed Likert scale | √ | √ | | | | |
| **Liu et al[135]** | clinical decision support(SDS) | Author-designed | Cross-sectional comparative study | Self-designed Likert scale | | | | | | usefulness, acceptance, relevance, understanding, workflow, bias, inversion, and redundancy,word count |
| **Munoz-Zuluaga et al[80]** | Laboratory Medicine | Textbook\professional insitustion\ clinical cases | Non-comparative study | Direct evaluation without scale | √ | √ | | | | |
| **Chervenak et al[115]** | fertility | Professional society and institutions | Non-comparative study | Self-designed Likert scale | √ | | | | | Word count, reference source |